\documentclass[pdflatex,sn-mathphys-num ]{sn-jnl}

\usepackage{graphicx}%
\usepackage{multirow}%
\usepackage{amsmath,amssymb,amsfonts}%
\usepackage{amsthm}%
\usepackage{mathrsfs}%
\usepackage[title]{appendix}%
\usepackage{xcolor}%
\usepackage{textcomp}%
\usepackage{manyfoot}%
\usepackage{booktabs}%
\usepackage{algorithm}%
\usepackage{algorithmicx}%
\usepackage{algpseudocode}%
\usepackage{listings}%
\usepackage{array}
\usepackage{graphicx}
\usepackage{subfig}%

\theoremstyle{thmstyleone}%
\theoremstyle{thmstyletwo}%

\theoremstyle{thmstylethree}%

\begin{document}


\title[Article Title]{AI-Enabled User-Specific Cyberbullying Severity Detection with Explainability }


\author*[1,2]{\fnm{ Tabia Tanzin Prama}}\email{tprama@uvm.edu}

\author[2]{\fnm{Jannatul Ferdaws Amrin}}\email{amrncse26@gmail.com@gmail.com}

\author[2]{\fnm{Md. Mushfique Anwar} }\email{manwar@juniv.edu}

\author[3]{\fnm{ Iqbal H. Sarker} }\email{m.sarker@ecu.edu.au }

\affil*[1]{\orgdiv{Computer Science}, \orgname{University of Vermont}, \orgaddress{\street{85 South Prospect Street}, \city{Burlington}, \postcode{05405}, \state{Vermont}, \country{United States}}}

\affil[2]{\orgdiv{Computer Science and Engineering}, \orgname{Jahangirnagar University}, \orgaddress{\city{Savar}, \postcode{1342}, \state{Dhaka}, \country{Bangladesh}}}

\affil[3]{\orgdiv{Centre for Securing Digital Futures}, \orgname{Edith Cowan University}, \orgaddress{\street{270 Joondalup Dr}, \city{Joondalup}, \postcode{6027}, \state{WA}, \country{Australia}}}


\abstract{The rise of social media has significantly increased the prevalence of cyberbullying (CB), posing serious risks to both mental and physical well-being. Effective detection systems are essential for mitigating its impact. While several machine learning (ML) models have been developed, few incorporate victims' psychological, demographic, and behavioral factors alongside bullying comments to assess severity. In this study, we propose an AI model intregrating user-specific attributes, including psychological factors (self-esteem, anxiety, depression), online behavior (internet usage, disciplinary history), and demographic attributes (race, gender, ethnicity), along with social media comments. Additionally, we introduce a re-labeling technique that categorizes social media comments into three severity levels: Not Bullying, Mild Bullying, and Severe Bullying, considering user-specific factors. Our LSTM model is trained using 146 features, incorporating emotional, topical, and  word2vec representations of social media comments as well as user-level attributes and it outperforms existing baseline models, achieving the highest accuracy of 98\% and an F1-score of 0.97. To identify key factors influencing the severity of cyberbullying, we employ explainable AI techniques (SHAP and LIME) to interpret the model's decision-making process. Our findings reveal that, beyond hate comments, victims belonging to specific racial and gender groups are more frequently targeted and exhibit higher incidences of depression, disciplinary issues, and low self-esteem. Additionally, individuals with a prior history of bullying are at a greater risk of becoming victims of cyberbullying.  }




\maketitle

\section{Introduction}

With the rapid advancement of technology and digitization, online communication has become an integral part of daily life. Social media platforms, messaging services, and gaming networks facilitate interactions between individuals across different cultures and backgrounds. However, these digital platforms have also given rise to cyberbullying, which involves using online forums, text messages, and emails to harass, threaten, or harm others. Cyberbullying can take various forms, such as making offensive comments, spreading false information, sharing embarrassing images or videos, and sending harmful messages (UNICEF, \cite{1}). According to Hinduja \cite{3}, cyberbullying involves repeated or publicly distributed acts of cruelty using electronic communication tools against individuals who are unable to defend themselves.

Cyberbullying is a growing global concern, particularly affecting children and adolescents. Unlike traditional bullying, cyberbullying is persistent, anonymous, and pervasive, making it difficult for victims to escape its impact. Research from the Cyberbullying Research Center indicates that over 34\% of students in the United States have experienced cyberbullying, with 17\% reporting incidents in the past 30 days \cite{3}. A large-scale EU youth internet study (2014) \cite{4} found that 20\% of 11- to 16-year-olds had encountered hateful content online, with exposure to cyberbullying increasing by 12\% from 2010 to 2014. Furthermore, studies suggest that cyberbullying disproportionately affects girls more than boys and is more common among individuals considered vulnerable or different.

The anonymity and distance provided by digital platforms often embolden offenders to act aggressively, resulting in severe consequences for victims. Cyberbullying has been linked to anxiety, depression, emotional distress, and even suicide. Notable cases include Phoebe Prince, a 15-year-old girl who took her own life after facing harassment on Facebook \cite{7}, a 14-year-old girl who died by suicide after receiving hateful comments on Ask.fm. A report found that nine suicides have been linked to cyberbullying on Ask.fm alone \cite{8}, highlighting the urgency of implementing effective cyberbullying detection and prevention strategies.

Despite increasing efforts, one of the major challenges in combating cyberbullying is the lack of accessible datasets for research and analysis. Many studies rely on manually curated datasets from social media platforms, making it difficult to generalize findings and develop scalable solutions \cite{10}. Some of the most prominent social media platforms have been identified as hubs for cyberbullying, with a survey by Ditch the Label Anti-Bullying Charity ranking Instagram (42\%) and Facebook (37\%) as the most reported platforms for online harassment \cite{9}. Given the growing prevalence of cyberbullying, automated detection and intervention systems are necessary to ensure a safer digital environment.

The Role of Explainable AI (XAI) \cite{sarker2024ai} in Cyberbullying Detection
Artificial Intelligence (AI) and Machine Learning (ML) have been increasingly utilized to detect cyberbullying by analyzing online interactions. However, most AI models, particularly deep learning networks, operate as black boxes, making their decision-making processes difficult to interpret. XAI addresses this issue by enhancing transparency, interpretability, and accountability in AI models, thereby increasing trust in automated systems (DARPA XAI Program) \cite{13}. XAI techniques such as LIME (Local Interpretable Model-Agnostic Explanations) and SHAP (SHapley Additive exPlanations) enable users to understand the factors influencing AI-driven cyberbullying detection.

Psychological learning also plays a crucial role in cyberbullying research. Understanding the cognitive and emotional processes behind cyberbullying behavior can help improve detection models. Recent studies combining qualitative and quantitative methods have identified social norms, emotional regulation, and perceived support as significant predictors of cyberbullying behavior \cite{14}. However, there is a gap in research integrating psychological learning theories with Explainable AI models to enhance the interpretability of cyberbullying detection systems. This study aims to develop an Explainable AI-driven cyberbullying detection model that integrates psychological learning and deep learning techniques to improve detection accuracy and interpretability. The key objectives are:

\begin{itemize}
    \item First, we determine whether a user is experiencing cyberbullying by analyzing their user profile information and a collection of received comments.
    \item The raw data undergoes preprocessing, including cleansing, followed by processing through the NLTK libraries' pipeline, which involves normalization, tokenization, stop word removal, stemming, and lemmatization.
    \item Feature engineering is performed using emotional features, topic modeling, online behavior features, user-level features, and word2vec-based feature representation.
    \item To enhance classification efficiency, we employ a deep learning model with Long Short-Term Memory (LSTM) using 146 features.
    \item and Lastly we use explainable AI tools (SHAP and LIME)to interpret the model’s predictions and factors behind the severity of cyberbullying and also compare with other existing baseline models.
\end{itemize}

By integrating Explainable AI with psychological learning principles, this research seeks to improve cyberbullying detection models, ensuring that AI-driven systems are interpretable, reliable, and actionable. The findings of this study can contribute to the development of more ethical and effective AI-based moderation strategies, fostering safer digital environments.

\section{Literature Review}
The automatic detection of hate speech has been a major research focus in the fields of data mining, natural language processing (NLP), and machine learning. The growing influence of social media platforms has intensified research interest in this domain. Various approaches, including machine learning, deep learning, and explainable AI (XAI), have been explored for hate speech detection.

The detection of cyberbullying and hate speech is a complex yet essential task due to the widespread impact of social networks on individuals, particularly adolescents. One study \cite{15} conducted a systematic review of 22 studies on cyberbullying detection and evaluated methodologies using two datasets. The findings revealed inconsistencies in defining cyberbullying, leading to inaccurate models with limited real-world applicability. The study emphasized the need for standardizing definitions and methodologies to enhance detection effectiveness. Traditional cyberbullying detection methods have primarily focused on analyzing individual comments without considering the relational dynamics between them. A novel approach proposed in \cite{16} models user interactions, taking into account topic coherence and temporal dynamics, which significantly improves detection accuracy. This highlights the necessity of incorporating contextual and behavioral patterns in cyberbullying detection systems.

While the application of stacking classifiers has been explored in various domains, including diabetes prediction \cite{17} and indoor scene classification \cite{18}, their use in hate speech detection is still evolving. Studies have demonstrated that ensemble learning techniques enhance classification accuracy by leveraging multiple classifiers \cite{19}. For instance, sentiment analysis using Naive Bayes, Support Vector Machine (SVM), and ensemble methods has been successfully applied to movie reviews [19]. Similarly, Turkish text classification has shown promising results using Naive Bayes models, achieving a 90\% classification success rate \cite{20}.The use of deep learning models, particularly Convolutional Neural Networks (CNNs), has shown improvements over traditional machine learning classifiers. Studies have found CNN-based models to be effective in processing large textual datasets, including identifying harmful online comments \cite{23}, \cite{28}. CNNs offer advantages such as automatic feature extraction and improved classification accuracy compared to manual feature engineering approaches.

Hybrid models, such as CNN-LSTM architectures, have also been explored to leverage both spatial and sequential text features. A study \cite{34} demonstrated that integrating CNN for regional feature extraction with LSTM for contextual analysis resulted in improved classification performance. Moreover, the use of bidirectional LSTM models in NLP tasks has been shown to enhance accuracy by capturing dependencies across text sequences \cite{31}, \cite{32}. In another study \cite{30}, a deep learning-based text classification model was proposed for hate speech detection on Twitter. The classifier categorized tweets into non-hate speech, racism, sexism, or both. The best-performing model, trained on word2vec embeddings, achieved an F-score of 78.3\%. The research highlights the significance of embedding-based representations in improving hate speech classification. Further advancements have been made in leveraging explainability techniques to understand deep learning decisions. The NA-CNN-LSTM model \cite{32} eliminates the activation function in CNNs to improve classification. Another study \cite{33} introduced a Double Channel (DC) technique, incorporating both word-level and char-level embeddings to enhance generalization, outperforming standard CNN-LSTM models.

XAI plays a crucial role in hate speech detection by providing transparency and interpretability to model predictions. Traditional AI models often lack explainability, making them challenging to deploy in sensitive applications such as content moderation and law enforcement. Research has shown that XAI techniques, such as Local Interpretable Model-Agnostic Explanations (LIME) and Shapley Additive Explanations (SHAP), can be integrated with deep learning models to provide interpretable insights \cite{38}. A recent study \cite{39} demonstrated the effectiveness of XAI-based approaches in detecting hate speech by using LIME and SHAP on a dataset of offensive tweets. The study achieved an accuracy of 97.6\% using an LSTM model and explored BERT-based variants to improve explainability. Another work \cite{40} compared XGBoost models with LSTM-based black-box models, showing that XGBoost models, combined with SHAP, provided better interpretability while maintaining high classification accuracy. Moreover, research in XAI has extended to multimedia applications, including text, images, audio, and video \cite{41}. The need for explainable hate speech detection models is particularly important in automated content moderation, where black-box decisions may lead to biased or unjustified content removals. Studies have proposed various generative enhancement techniques to improve model interpretability and mitigate risks associated with biased predictions \cite{26}. Interactive tools, such as XNLP, have been developed to provide real-time insights into explainability research in NLP \cite{42}. These tools facilitate knowledge sharing and exploration of XAI methodologies, contributing to the broader adoption of explainable models. Additionally, XAI has been applied in specific domains, such as xenophobia detection in online communities, where interpretable models help decision-makers mitigate hate speech-driven violence \cite{43}.

Recent advancements in deep learning and XAI have significantly improved the accuracy and interpretability of hate speech classification models. However, challenges remain in areas such as data imbalance, contextual understanding, and cross-platform generalization. Future research should focus on developing robust models that address biases, incorporate multimodal features, and enhance explainability for real-world deployment \cite{44}. Moreover, the integration of decentralized deep learning approaches, such as MaLang \cite{35}, has shown promise in enhancing cybersecurity by preventing the transmission of harmful content across organizational networks. Similarly, MetaHate \cite{36} has demonstrated improvements in hate speech detection by combining multiple predictors, further reinforcing the importance of ensemble approaches in NLP. As AI continues to evolve, it is essential to balance model accuracy with explainability to ensure fair and ethical content moderation. The development of hybrid models that combine deep learning with explainability techniques will play a crucial role in shaping the future of hate speech detection.

\section{Methodology}
\begin{figure}[H]
    \centering
    \includegraphics[width=0.7\linewidth]{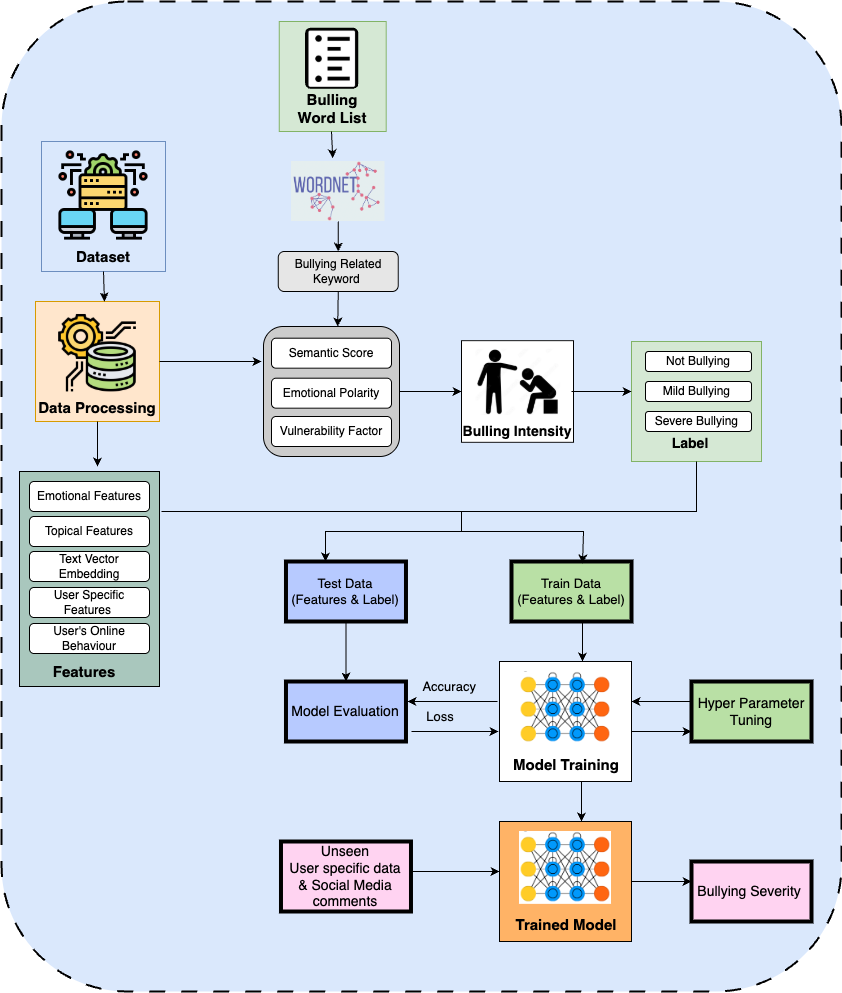}
    \caption{Overview of Proposed Workflow  to determine intensity of bullying and classify cyberbullying instances into ``Not bullying", ``Mild", and ``Severe" categories }
    \label{fig:workflow}
\end{figure}

An LSTM deep learning model has been adopted to classify cyberbullying victims using user data, vulnerability characteristics, and comments on social media.Figure \ref{fig:workflow} shows the workflow of the proposed method to classify cyberbullying instances into ``Not bullying", ``Mild", and ``Severe" categories. The dataset, sourced from BullyBlock \cite{50}, undergoes preprocessing, including data cleaning, tokenization, stemming, and word embedding. Emotional polarity and vulnerability factors are assessed to determine bullying intensity. Five distinct features are extracted and used to train the LSTM classifier, which processes textual and numerical features in parallel to detect harmful remarks and predict toxicity levels. And lastly the proposed model is extensively evaluated and benchmarked against multiple state-of-the-art models for performance comparison.

\subsection{Dataset}
In this study, we utilized the BullyBlocker dataset \cite{50}, which comprises 400 adolescents and contains 22,915 positive samples (instances of cyberbullying) and 17,439 negative samples (normal records). Each record consists of two main components: user profile information (including user ID and vulnerability-related factors) and the set of messages received by the user. The dataset includes details such as sender ID, timestamp, and message content. To identify cyberbullying instances, we defined a 90-day (D) limit and a maximum of 100 messages (M) per user. The dataset was obtained through login information from Facebook, capturing wall posts, comments, and user attributes such as age, gender, ethnicity, race, frequency of daily internet use, and vulnerability-related factors (e.g., depression, self-esteem issues, and disciplinary history). The bullying history attribute was distributed as follows: no previous bullying (50\%), experienced bullying last month (16.66\%), between one to two months ago (16.66\%), and more than two months ago (16.66\%). This dataset provided a valuable resource for analyzing the relationship between vulnerability factors and cyberbullying risk. A sample dataset, including user information, vulnerability factors, and received comments, is presented in Appendix \ref{appendix-table}. Using profile information and received messages, we determined whether a user was experiencing cyberbullying by assigning a Bullying Rank score (ranging from 0 to 100), representing the likelihood that the user had recently faced or was currently facing cyberbullying.

\subsection{Data Preprocessing}
To prepare the raw dataset for analysis, we applied data cleaning, normalization, and preprocessing filters. This process involved removing unwanted data tokens, such as HTML elements, extra white spaces, special symbols, and numerals. Missing data was handled by eliminating instances with more than 50\% missing values, while remaining empty cells were filled using either the mean of related values or the minimum value to maintain dataset integrity. Additionally, duplicate entries, including posts and tweets, were removed to avoid redundancy.

\begin{figure}[H]
    \centering
    \includegraphics[width=0.6\linewidth]{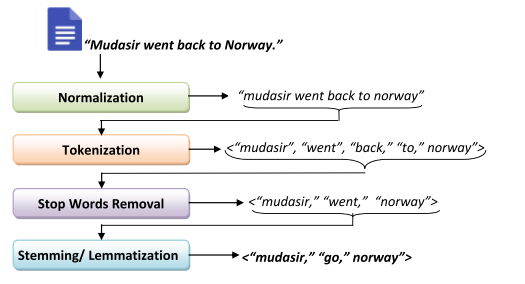}
    \caption{ Data preprocessing using NLTK libraries.
}
    \label{fig:pre-process}
\end{figure}

Following data cleaning, we processed the dataset using the Natural Language Toolkit (NLTK) for text preprocessing and analysis. The NLTK pipeline, as illustrated in Figure \ref{fig:pre-process}, included normalization, tokenization, stop word removal, stemming, and lemmatization. Tokenization segmented the text into words and phrases (tokens), aiding in context understanding and NLP model construction. Normalization ensured uniformity by converting all text to lowercase. Stop words, such as ``the," ``is," and ``and," which provide minimal value for classification, were removed before model training. Stemming reduced words to their root forms, allowing different word variations to be mapped to a common stem, even if the stem itself was not a valid word in the language \cite{50}.

\subsection{Bullying Intensity Measurement and Data Labeling}

For the given dataset, we developed a bullying intensity measurement technique inspired by the approach used for depression intensity measurements \cite{10510456}. The bullying score is computed based on three key components: emotional polarity of comments, user vulnerability factors, and semantic similarity. And Then the bullying intensity is categorized into three predefined levels: ``Not Bullying" \([0, 0.33]\), ``Mild Bulling" \([0.34, 0.66]\), and ``Severe Bulling" \([0.67, 1]\).  

The emotional polarity of comments is assessed using the NLTK \footnote{https://www.nltk.org/} library, which analyzes the negative polarity of user posts and assigns a polarity score in the range \([-1,1]\). Sentiment classification is performed using VADER (Valence Aware Dictionary and Sentiment Reasoner)\footnote{https://github.com/cjhutto/vaderSentiment}, producing sentiment scores in four categories: negative (neg), neutral (neu), positive (pos), and compound (compound). The compound score, a normalized sentiment score within the range \([-1,1]\), is used as an indicator of bullying intensity. Higher negative polarity values contribute positively to the overall bullying score.  

The vulnerability factor (VF) measures how susceptible a monitored adolescent is to cyberbullying. This is determined by several demographic and behavioral attributes, including age, gender, race and ethnicity, prior history of bullying, frequency of internet use, internalizing problems (mental health history), and externalizing problems (past disciplinary issues). Each vulnerability factor is assigned a weight within the range \([0,1]\), based on association coefficients from prior meta-analytic studies. The weighting is derived from correlation coefficients reported by Kowalski et al. \cite{62} and Guo \cite{63} for age, bullying history, internalizing disorders, and externalizing issues. The weight for internet usage frequency follows Kowalski et al. \cite{62}, while gender and race/ethnicity weights are based on Guo \cite{63}. The overall vulnerability score is computed by normalizing each vulnerability factor by its assigned weight.Each vulnerability factor and its corresponding weight are detailed in Table \ref{tab:vulnerability_factors}

\begin{table}[ht]
\centering
\begin{tabular}{|>{\raggedright}m{0.25\linewidth}|>{\raggedright}m{0.55\linewidth}|>{\centering\arraybackslash}m{0.1\linewidth}|}
\hline
\textbf{Vulnerability Factor} & \textbf{Details} & \textbf{Weight} \\
\hline
Age & 11-16 Years & 0.04 \\
\hline
Gender & female & 0.12 \\
\hline
Race/Ethnicity & race is nonwhite \newline ethnicity is Hispanic/Latino & 0.02 \\
\hline
Past Bullying & user experienced bullying in last 1 month, \newline 1-2 months, more than 2 months & 0.42 \\
\hline
Daily Internet Use & 4h-6h, \newline $>$6h & 0.17 \\
\hline
Internal Issues & Having depression, low self-esteem, anxiety & 0.28 \\
\hline
External Issues & Having disciplinary issues or substance use & 0.21 \\
\hline
\end{tabular}
\caption{Users vulnerability factors and assigned weight for data labeling}
\label{tab:vulnerability_factors}
\end{table}

To collect bullying-related seed terms, we gathered adolescents’ opinions on common forms of bullying they experienced on social media, as shown in Table \ref{tab:bullying_keywords}. We then expanded these keywords using Latent Semantic Indexing (LSI)\footnote{https://nlp.stanford.edu/IR-book/html/htmledition/latent-semantic-indexing-1.html}, which applies Singular Value Decomposition (SVD) to uncover patterns in text. LSI connects semantically similar words based on their usage context. Finally, we calculated a semantic score for each comment by computing the weighted sum of normalized term hits and scaling it to [0,1].

\begin{table}[h]
    \centering
    \renewcommand{\arraystretch}{1.5}
    \begin{tabular}{|p{7cm}|}
        \hline
        \multicolumn{1}{|c|}{\textbf{Bullying Keywords}} \\ 
        \hline
        Ahole, amcik, anal, analprobe, andskota, anilingus, bukkake, bull shit, bullshit, 
        bullshits, bullshitted, bullturds, bung, busty, butt, butt fuck, butt-pirate, 
        buttfuck, buttfucker, chode, chodes, chraa, chuj, C11t, climax, clit, clitoris, 
        clitorus, clits, clitty, flikker, flipping the bird, floozy, foad, fondle, foobar, 
        foreskin, fotze, freex, frigg, frigga, fubar, fuck, packie, packy, paddy, paki, 
        pakie, paky, pantie, panties, panty, paska, fuckwhit, fuckwit, etc. \\
        \hline
    \end{tabular}
    \caption{Example of frequently used bullying words }
    \label{tab:bullying_keywords}
\end{table}

\begin{equation}
S_{\text{total}} = \text{PS} + \text{SS} + \text{VF}
\label{eq1}
\end{equation}

\begin{equation}
\text{Bullying Intensity (BL)} = \frac{S_{\text{total}} - \min}{\max - \min}
\end{equation}

\begin{equation}
\text{Label} =
\begin{cases} 
\text{Not Bullying}, & 0 \leq \text{BI} \leq 0.33 \\
\text{Mild Bullying}, & 0.34 \leq \text{BI} \leq 0.66 \\
\text{Severe Bullying}, & 0.67 \leq \text{BI} \leq 1
\end{cases}
\label{class}
\end{equation}

First, we calculate the total bullying score by summing the polarity score \textit{(PS)}, vulnerability factor \textit{(VF)}, and semantic score \textit{(SS)} using Equation \ref{eq1}. We then apply min-max normalization, where \textit{min} and \textit{max} represent the minimum and maximum values of the bullying score \( S_{\text{total}} \). This ensures that the user bullying intensity \textit{(BI)} falls within the range \([0,1]\). 

Finally, using Equation \ref{class}, we categorize the severity of bullying based on the bullying intensity score. Table \ref{tab:sample_comments} presents examples of comments along with their corresponding bullying intensity scores and assigned class labels.

\begin{table}[ht]
\centering
\begin{tabular}{|>{\raggedright}m{0.4\linewidth}|>{\centering}m{0.15\linewidth}|>{\centering\arraybackslash}m{0.25\linewidth}|}
\hline
\textbf{Comment} & \textbf{Bullying Intensity} & \textbf{Assigned Class Label} \\
\hline
lmao how tf is this a costume for \$30 u can b a dumbass bitch and get ya ass beat by me fuck spirit https://t.co/42jcxs1q7m & 0.84 & Severe Bullying \\
\hline
looking for a car in a car with no a/c has me wanting 2 die & 0.23 & Not Bullying \\
\hline
"i know i'm probably some bitch to you" & 0.54 & Mild Bullying \\
\hline
bro my sister been sleep since like 3 like uh bitch is you pregnant. & 0.62 & Mild Bullying \\
\hline
this bitch looks like a selena wannabe & 0.69 & Severe Bullying \\
\hline
what i wouldn't give to have obama back again. he was a leader. trump is a loser, liar, fool, and traitor. & 0.72 & Severe Bullying \\
\hline
trackering app is real you can't play with people live anymore & 0.11 & Not Bullying \\
\hline
\end{tabular}
\caption{Example of Sample comments, bullying intensity score, and assigned class label}
\label{tab:sample_comments}
\end{table}

\subsection{ Feature Engineering
}

Our objective is to detect and analyze cyberbullies by examining online behaviors. Key features such as emotion, events, user-specific characteristics, online behavior, and bullying comments play a crucial role in the detection of cyberbullying. By analyzing these factors, we aim to accurately identify cyberbullies and implement strategies to prevent and mitigate the impact of cyberbullying.

\subsubsection{Emotional Features}In order to detect cyber bullies from their online behaviors, we analyze various features, including emotional features. We specifically consider 9 dimensions of emotion-related features, which are crucial for identifying negative sentiment in social media posts. To achieve this, we rely on tools such as Linguistic Inquiry and Word Count (LIWC) to count the number of negative words in the tweets \cite{53}. Additionally, we extract emotional features at both the sentence level (4-D) and word level (4-D) to capture the nuances of the language used in social media. These features play an essential role in detecting cyberbullying and help us develop effective strategies for preventing such behavior.
\subsubsection{Topical/Event Features:
}

In computational linguistics, topic modeling is an effective technique for narrowing the feature space of textual data to a set of themes \cite{54}. Unsupervised text mining extracts latent topics from documents, including those related to cyberbullying. Unlike LIWC, which uses a fixed word set, LDA generates unlabeled word groups based on likelihood. This study analyzed semantic connections in cyberbullying-related Twitter posts using LDA to determine topic distributions. As a probabilistic generative model, LDA uncovers underlying topic structures in discrete data [54]. Limiting LDA to 25 topics improved performance on the validation set, considering only terms appearing in 10+ posts. Each post, treated as a separate document, was tokenized and stemmed before topic calculation. Stop words were removed before modeling, and Mallet toolbox was used for LDA implementation \cite{55}.

\subsubsection{User Level Feature
}
We incorporate user-level features to identify potential cyberbullies. The following user attributes are considered: ``id," ``Age", ``Gender", ``Ethnicity", ``Race", ``Depression", ``Anxiety", ``SelfEsteemIssues", ``DisciplinaryIssues", ``SubstanceAbuse", ``Date", etc. To process these attributes, we remove dates, Names, and IDs. For attributes that have boolean values, we convert them to binary form (0/1). For attributes with unique values, we obtain the unique sets. This enables us to identify key demographic information and individual characteristics that may be associated with cyberbullying behavior.

\subsubsection{ Word Vector Representation
}

Word2Vec is a key tool for analyzing word relationships in a corpus. It employs neural network models to generate word embeddings, positioning similar words closely and dissimilar words far apart in vector space \cite{56}. This technique effectively identifies hidden word correlations and patterns.

Word2Vec utilizes Continuous Bag of Words (CBoW) and skip-gram training methods. The CBoW model predicts a target word based on surrounding context, while skip-gram predicts context words from a target word \cite{57,58}. The objective of CBoW is to maximize the log probability of predicting the target word \( w_t \) given its surrounding words:

\begin{equation}
\sum \log p(w_t | w_{t-N}, ..., w_{t-1}, w_{t+1}, ..., w_{t+N})
\end{equation}

where \( p(w_t | w_{t-N}, ..., w_{t+N}) \) represents the probability of predicting \( w_t \) given its context window of size \( N \). The output of Word2Vec consists of word vectors, where semantically similar words (e.g., sobbing and pain) are closely placed, whereas unrelated words (e.g., green and nature) are far apart.

\begin{figure}[h]
    \centering
    \includegraphics[width=0.3\textwidth]{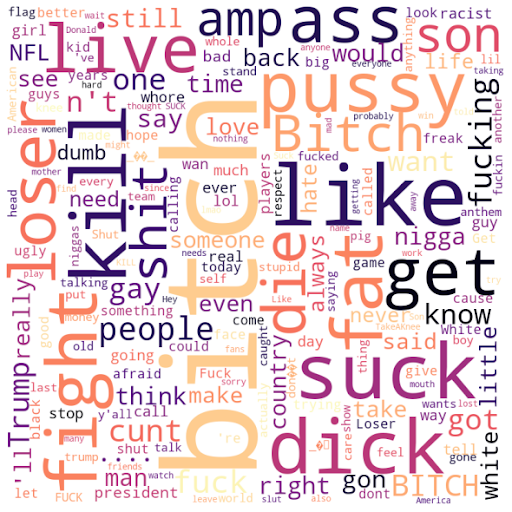}
    \caption{Word cloud of frequently used in bullying comments.}
    \label{fig:wordcloud}
\end{figure}

We trained Word2Vec on our dataset using the CBoW model to extract vocabulary words. The word cloud in Figure~\ref{fig:wordcloud} highlights the most frequently used words in bullying-related content, including ``bitch" (5467 occurrences), ``fuck" (1231), ``suck" (1169), ``kill" (1146), ``dick" (1113), ``fat" (969), ``fight" (961), ``pussy" (950), ``loser" (794), and ``shit" (670).

 In total, 146 feature vectors were utilized, comprising emotional features (12 vectors), topic-level features (25 vectors), user-level features (9 vectors), and the 100-dimensional Word2Vec representation. These feature vectors encompassed a diverse range of elements essential for accurate model training and analysis.

\section{Background Study}

\subsection{Proposed LSTM-based Model
}

\begin{figure}
    \centering
    \includegraphics[width=0.7\linewidth]{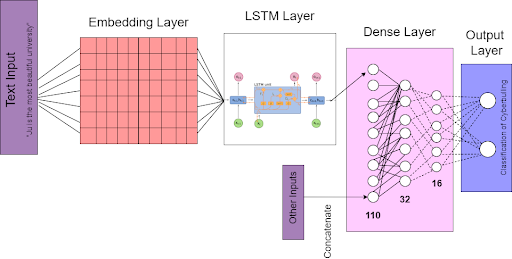}
    \caption{Architecture of the proposed LSTM model for cyber bullying severity prediction}
    \label{fig:model}
\end{figure}

In this study, we propose a Long Short-Term Memory (LSTM)-based deep learning model for cyberbullying detection to enhance classification performance. The architecture of the proposed model, designed to effectively capture temporal relationships and semantic meanings in textual data, is illustrated in Figure \ref{fig:model}. The model follows a sequential architecture with multiple layers, including two input layers, an embedding layer, an LSTM layer, a concatenation layer, two dense layers, and an output layer. The first input layer processes textual data with a maximum input length of 100, while the second input layer accommodates additional non-textual features (emotional, topic-level, \& user-level) with a shape of 46. The embedding layer converts input text into a dense vector representation to capture contextual meanings. We utilize a pre-trained embedding matrix and configure the embedding layer to be trainable. Our dataset analysis indicates an average comment length of 15.09, a maximum length of 277, and a standard deviation of 35.34. The vocabulary size is 37,201. To ensure uniform input length, we apply padding and truncation using the Keras `pad\_sequence` function, setting \textit{maxlen} = 100, \textit{padding} = 'post', and \textit{truncating} = 'post'. The embedding layer is defined with an embedding size of 37,202, an embedding dimension of 100, an input length of 100, and \textit{trainable} = True. The \textit{LSTM layer} is the core component of the model, responsible for analyzing text input and extracting critical features for classification. It consists of 64 memory units, which regulate information flow through a series of gates, enabling the model to retain and utilize relevant sequential dependencies. The LSTM output is then combined with non-textual feature input using a \textit{concatenation layer}. Following the concatenation, the model includes two fully connected dense layers with 32 and 16 neurons, respectively, both employing the \textit{ReLU} activation function. Finally, the \textit{output layer} consists of a single dense neuron with a \textit{sigmoid} activation function, which outputs the probability of the input text belonging to a cyberbullying class.

\subsection{Baseline Models}

\subsubsection{Logistic regression (LR)}
LR is a linear classification algorithm that models the probability of a sample belonging to a specific sentiment class. It applies the sigmoid activation function to map predicted values between 0 and 1, making it effective for binary and multi-class classification \cite{Das2014}. Logistic Regression is well-suited for sentiment analysis tasks, especially in real-time applications like social media monitoring and customer feedback analysis, due to its interpretability and low computational complexity.

\subsubsection{Support Vector Machine (SVM)} SVM is a powerful classification algorithm that finds an optimal hyperplane to separate sentiment classes while maximizing the margin between them \cite{cortes1995support}. It efficiently handles high-dimensional and sparse text data, making it suitable for bag-of-words and TF-IDF representations. The kernel trick allows SVM to model non-linear relationships, making it highly effective for complex sentiment classification.

\subsubsection{Random Forest (RF)} RF is an ensemble learning algorithm that builds multiple decision trees using bootstrap sampling and aggregates their outputs \cite{ho1995random}. It enhances model robustness by reducing overfitting, making it well-suited for handling noisy, incomplete, and high-dimensional text data. It also provides feature importance ranking, making it valuable for understanding key sentiment indicators.

\subsubsection{Deep neural network (DNN)} DNNs are multi-layer perceptron (MLP)-based deep learning architectures that model non-linear relationships between text features and sentiment. A DNN consists of an input layer, multiple hidden layers with activation functions and an output layer for classification. It excels at capturing subtle sentiment variations and complex linguistic patterns. For comparison, we replicate the proposed LSTM model architecture, replacing LSTM layers with DNN layers while maintaining all other components.

\subsubsection{GRU} GRU is a variant of Recurrent Neural Networks (RNNs) that mitigates the vanishing gradient problem using gating mechanisms. GRUs capture long-term dependencies in sequential text data, making them ideal for sentiment analysis in long texts, customer reviews, and social media posts. Unlike traditional RNNs, GRUs use update and reset gates to control information flow. Similar to the DNN model, we replicate the proposed architecture with GRU layers for the GRU model.

\subsubsection{CNN} CNNs apply convolutional filters to learn hierarchical text representations, detecting local patterns such as n-grams in text data. With multiple convolutional and pooling layers, CNNs capture both regional and global sentiment features, making them well-suited for large-scale sentiment analysis tasks input data and manage enormous datasets. Here we also follow the same architecture of proposed model.

\subsection{Explainable Artificial Intelligence (XAI)
}

 XAI enhances the transparency, fairness, and interpretability of machine learning models, addressing the challenges of AI decision-making and bias detection \cite{64}. As AI models grow more complex, they often function as "black boxes," making it difficult for even developers to understand their decision-making processes \cite{65}. To overcome this, we employ LIME (Locally Interpretable Model-Agnostic Explanations) and SHAP (Shapley Additive Explanations), which provide insights into how models generate predictions and how different features influence these outcomes \cite{66}.

\begin{itemize}
    \item \textbf{Locally Interpreted Model-Agnostic Explanations (LIME).}
LIME approximates complex model predictions by creating an interpretable local model that explains individual decisions. It perturbs input values and observes changes in output, identifying feature importance for each prediction. The generated explanations aid in understanding model behavior, and we include LIME results for all machine learning models in our study \cite{68}.

\item \textbf{Shapley Additive Explanations (SHAP).}
SHAP, based on game theory’s Shapley values, quantifies each feature’s contribution to the final prediction \cite{69,70}. Unlike traditional feature importance methods, SHAP ensures consistency and can be applied to deep learning, decision trees, and linear models \cite{71}. By averaging contributions across feature subsets, SHAP provides both global and local explanations, offering a more precise interpretation of model predictions.
\end{itemize}

\section{Result and Analysis
}

\subsection{Performance Evaluation}

The performance of our proposed LSTM-based cyberbullying detection model was evaluated using key metrics, including accuracy, recall, precision, and F1-score.Table \ref{tab:performance_evaluation} shows, with a comprehensive feature set (Emotion + Topic + Online Behaviour + User Level + Word2Vec) trained on the LSTM network, our model achieved 98\% accuracy, 0.97 recall, 0.98 precision, and an F1-score of 0.97, highlighting its effectiveness in identifying instances of cyberbullying. To analyze feature contributions, we trained the model on individual feature subsets, where the Word2Vec representation of comments emerged as the most significant, achieving 77\% accuracy, 0.82 recall, 0.84 precision, and an F1-score of 0.83, reinforcing the importance of comment-based textual analysis in cyberbullying detection. Emotional features, due to their variability, contributed minimally to classification accuracy. To enhance model performance at the user level, we integrated feature sets with user-specific attributes, where the "User level + Word2Vec" feature set outperformed others, demonstrating the significance of user context in cyberbullying detection. 

Considering the complete feature set (Emotion, Topic, Online Behaviour, User Level, and Word2Vec), we compare the performance of the proposed LSTM model with baseline models, as presented in Table \ref{tab:performance_evaluation22}. Random Forest exhibited strong performance with 90\% accuracy, 0.908 recall, 0.894 precision, and an F1-score of 0.90, while SVM achieved 84\% accuracy, 0.842 recall, 0.843 precision, and an F1-score of 0.711. Among deep learning models, we replicate the proposed architecture, replacing the LSTM layers with respective layers (CNN, DNN, GRU).
CNN outperformed DNN and GRU, attaining 87\% accuracy, 0.863 recall, 0.889 precision, and an F1-score of 0.875, while DNN achieved 84\% accuracy, 0.814 recall, 0.848 precision, and an F1-score of 0.821. Overall, our proposed LSTM-based model demonstrates exceptional performance, particularly with the full feature set, where Word2Vec with user-level features significantly enhances cyberbullying detection. Notably, Random Forest and CNN perform best in their respective categories, validating the effectiveness of our feature-rich deep learning-driven approach.

\begin{table}[ht]
\centering
\begin{tabular}{|>{\raggedright}m{0.20\linewidth}|>{\centering}m{0.12\linewidth}|>{\centering}m{0.12\linewidth}|>{\centering}m{0.12\linewidth}|>{\centering\arraybackslash}m{0.12\linewidth}|}
\hline
\textbf{Features} & \textbf{Accuracy} & \textbf{Precision} & \textbf{Recall} & \textbf{F1 Score} \\
\hline
Emotion & 65\% & 0.69 & 0.56 & 0.62 \\
\hline
Topic & 72\% & 0.74 & 0.70 & 0.72 \\
\hline
Online Behaviour & 68\% & 0.83 & 0.87 & 0.85 \\
\hline
User Level & 69\% & 0.95 & 0.69 & 0.80 \\
\hline
Word2Vec & 77\% & 0.82 & 0.84 & 0.83 \\
\hline
User Level + Word2Vec & 89\% & 0.89 & 0.62 & 0.89 \\
\hline
User Level + Emotion & 80\% & 0.81 & 0.78 & 0.79 \\
\hline
User Level + Topic & 88\% & 0.88 & 0.81 & 0.84 \\
\hline
User Level + Online Behaviour & 79\% & 0.81 & 0.78 & 0.79 \\
\hline
\textbf{Emotion + Topic + Online Behaviour + User Level + Word2Vec} & \textbf{98\%} & \textbf{0.97} & \textbf{0.98} & \textbf{0.97} \\
\hline
\end{tabular}
\caption{Performance Evaluation of Different Feature Sets in the Proposed Model}
\label{tab:performance_evaluation22}
\end{table}

\begin{table}[ht]
\centering
\begin{tabular}{|>{\raggedright}m{0.15\linewidth}|>{\centering}m{0.12\linewidth}|>{\centering}m{0.12\linewidth}|>{\centering}m{0.12\linewidth}|>{\centering\arraybackslash}m{0.1\linewidth}|}
\hline
\textbf{Model} & \textbf{Accuracy} & \textbf{Precision} & \textbf{Recall} & \textbf{F1 Score} \\

\hline
RF &  90\% & 0.894 & 0.908 & 0.90 \\
\hline

SVM &  84\% & 0.843 & 0.842 & 0.711 \\
\hline

\hline
LR &  72\% & 0.734 & 0.808 & 0.769 \\
\hline
DNN &  84\% &  0.848 &  0.814 
 & 0.821 \\
\hline
GRU & 74\% & 0.753 & 0.8107 & 0.780 \\
\hline
CNN &  87\% & 0.889   & 0.863 & 0.875 \\
\hline
\textbf{LSTM}  & \textbf{98\%} &\textbf{ 0.97} & \textbf{0.98} & \textbf{0.97} \\
\hline
\end{tabular}
\caption{Performance Evaluation Result of the Proposed Model}
\label{tab:performance_evaluation}
\end{table}

The results of our study, which are shown in Table \ref{tab:performance_evaluation}, shows that the proposed model is more powerful than earlier research in this field. With the help of important criteria including accuracy, recall, precision, and F1-score, we evaluated the performance of several models. All baseline models have been surpassed by our proposed model, incorporating a full feature set trained on the LSTM network. We attained impressive results with an accuracy of 98\%, recall of 0.97, precision of 0.98, and an F1-score of 0.97. These outcomes support the ability of our method to precisely identify instances of cyberbullying. In addition, we trained our model with feature subsets and performed a thorough analysis. With an accuracy of 77\%, recall of 0.82, a precision of 0.84, and an F1-score of 0.83, the word2vec representation of comments stood out as the most significant feature across these subgroups. This research emphasizes the value of looking through comment sections to detect instances of cyberbullying. However, because of their diversity, emotional features are  shown to be less illuminating and offer only minimal assistance in identifying, offer only minimal assistance in the identification of cyberbullying. We merged each feature set with user-specific attributes and trained an interconnected LSTM network to improve our model's performance to improve the performance of our model at the user level. In particular, the "User level + word2vec" feature produced better results than other features, highlighting the significance of taking user-level context into account in cyberbullying identification. 
The Random Forest classifier scored a remarkable accuracy of 90\%, recall of 0.908, a precision of 0.894, and an F1-score of 0.90 when using all five features when compared to the baseline models. With an accuracy of 84\%, recall of 0.842, precision of 0.843, and an F1-score of 0.711, the SVM model also performed admirably. Convolutional Neural Network (CNN) performed better than other deep learning models, including DNN and GRU, with an accuracy of 87\%, recall of 0.863, precision of 0.889, and an F1-score of 0.875. With an accuracy of 84\%, recall of 0.814, precision of 0.848, and an F1-score of 0.821, the DNN model produced encouraging results.
In conclusion, our proposed model exhibits remarkable performance in cyberbullying detection, especially when using the LSTM network with the full feature set. The ability to recognize instances of cyberbullying is greatly enhanced by the word2vec representation of comments, which also incorporates user-level information. Notably, the Random Forest classifier and CNN come out on top in their respective categories, reiterating the effectiveness of our strategy for identifying cyberbullying.

\subsection{ Explainabilty of the proposed LSTM model
}

For the explainability analysis, we categorize mild bullying and severe bullying as bullying to simplify visualization. In the first phase, we analyze bullying comments using the LIME XAI tool to identify the triggering words that influence the model’s prediction of bullying and non-bullying in cyberbullying detection. Table \ref{LIME1} presents examples of two test cases.

\begin{table}[ht]
\centering
\begin{tabular}{|>{\raggedright}m{0.05\linewidth}|>
{\raggedright\arraybackslash}m{0.2\linewidth}|>{\raggedright\arraybackslash}m{0.6\linewidth}|}
\hline
\textbf{No}  & \textbf{Class Label} & \textbf{Comments} \\
\hline
1& Bullying & let get son bitch white house take knee \\
\hline
2 & Not Bullying & nowplaying mylo feat miami sound machine docto... \\
\hline

\end{tabular}
\caption{  Two sample comments for analyzing explainability of the proposed model}
\label{LIME1}
\end{table}

\begin{figure*}[ht]
    \centering
    \subfloat[Explanation of Sample 1 in the Table \ref{LIME1}]{%
        \includegraphics[width=0.48\linewidth]{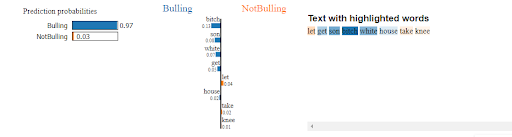}%
        \label{fig:LM2}}
    \hfill
    \subfloat[Explanation of Sample 2 in the Table \ref{LIME1}]{%
        \includegraphics[width=0.48\linewidth]{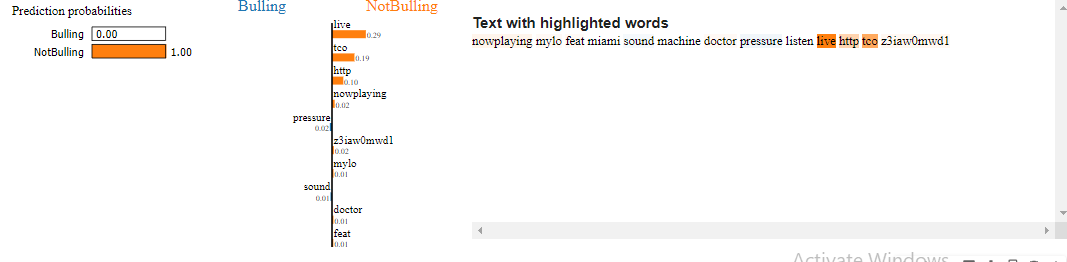}%
        \label{fig:LM3}}
    \caption{LIME explanations for Table \ref{LIME1} showing important words contributing to the model's decision.}
    \label{fig:LIME}
\end{figure*}

In Figure \ref{fig:LIME}, specific words or phrases contributing to the bullying classification are highlighted in blue, while those contributing to the non-bullying classification are highlighted in orange. For Sample 1 in Figure \ref{fig:LM2}, classified as bullying by our proposed model with a prediction probability of 0.97, the presence of negative words such as ``bitch" acted as a triggering factor, influencing the model’s decision to categorize the text as bullying. In Figure \ref{fig:LM3}, Sample 2 was classified as non-bullying with a prediction probability of 1.0. The presence of positive words like "live" and "nowplaying" contributed to the model’s decision to classify the text as non-bullying.

To explain the severity classification of cyberbullying victims beyond just analyzing bullying comments, we employ Shapley values to quantify the contribution of each feature to the model's prediction. Shapley values provide an attribution mechanism that assigns a credit score to each feature, indicating its influence on the model’s decision. These values help in interpreting the model’s output and understanding the relative importance of different factors in cyberbullying severity classification.

\begin{figure}[ht]
    \centering
    \includegraphics[width=0.6\linewidth]{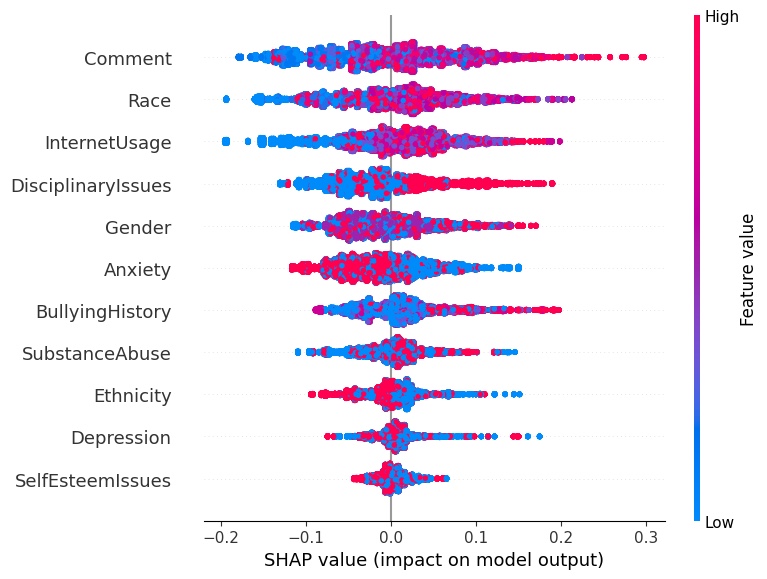}
    \caption{SHAP Summary Plot of the Proposed LSTM Model for Cyberbullying Severity Explanation. The positive SHAP values (towards the right) contribute to higher bullying severity classification, whereas the negative SHAP values (towards the left) contribute to lower bullying severity classification. Red dots represent high feature values, and blue dots represent low feature values.}
    \label{fig:shap1}
\end{figure}

As shown in Figure \ref{fig:shap1}, the features are ranked from most impactful at the top to least impactful at the bottom. The "Comment" feature, representing the textual content of the post, holds the greatest influence in determining the cyberbullying severity classification. However, other important factors, such as Race, Internet Usage, Disciplinary Issues, and Gender, also play a significant role, indicating that demographic and behavioral attributes contribute to the severity classification of cyberbullying victims. Individuals with a history of bullying others tend to have a higher likelihood of becoming victims of cyberbullying themselves. Additionally, disciplinary issues are strongly correlated with cyberbullying severity, where individuals with higher values (red dots) for this attribute are more likely to be classified under severe cyberbullying cases. Psychological attributes such as Anxiety and Depression also have a notable impact, as individuals with higher values in these attributes (red dots leaning right) are more susceptible to involvement in cyberbullying interactions, either as victims or perpetrators. Similarly, high internet usage emerges as a major contributing factor, as individuals who spend excessive time online are at a higher risk of encountering or engaging in cyberbullying. Demographic attributes such as Race, Gender, and Ethnicity also play a crucial role in understanding the severity of cyberbullying. A high feature value in Gender and Race suggests that individuals of certain genders or racial backgrounds are more likely to experience cyberbullying. This highlights the importance of considering socio-demographic factors when addressing cyberbullying prevention and intervention strategies.

\section{Conclusion
}

Adolescents are increasingly vulnerable to cyberbullying, which involves the intentional spread of false, humiliating, or hostile content through online platforms. In this study, we proposed an LSTM-based model to classify cyberbullying severity into three categories: ``Not Bullying", ``Mild Bullying", and ``Severe Bullying". The model incorporates user-specific attributes—such as psychological, demographic, and behavioral factors—alongside social media comments. Our results show that the proposed model outperforms baseline machine learning and deep learning models, achieving 98\% accuracy and an F1-score of 0.97, with Random Forest (RF) as the second-best performer. Incorporating user-specific factors significantly improved detection reliability, especially features like self-esteem issues, prior bullying history, and disciplinary records. 
XAI tools (SHAP and LIME) provided further insights, highlighting the influence of both textual and demographic attributes, including race and gender, on the classification outcomes. We deliberately excluded large pretrained language models like BERT or XLM-RoBERTa to maintain time efficiency and enable integration of user-specific data, which is challenging with such models. For future work, we aim to develop a browser extension for social networking platforms to alert guardians about potential cyberbullying risks in real time. We also plan to explore multimodal pretrained models like GPT or LLaMA for improved performance and extend our framework to detect cyberbullying in multimedia content such as images and videos.

\backmatter

\section*{Declarations}

\subsection*{Funding}  
The authors did not receive support from any organization for the submitted work. No funding was received to assist with the preparation of this manuscript.

\subsection*{Conflicts of Interest/Competing Interests}  
The authors have no relevant financial or non-financial interests to disclose. All authors certify that they have no affiliations with or involvement in any organization or entity with any financial interest or non-financial interest in the subject matter or materials discussed in this manuscript.

\subsection*{Authors’ Contributions}  
 Data collection, analysis, methodology and first draft of the manuscript was written by Tabia Tanzin Prama. Jannatul Ferdaws Amrin contributed to the study's literature review. Tabia Tanzin Prama , Md. Mushfique Anwar and Iqbal H. Sarker contributed to the study’s conception, design. All authors provided critical feedback and approved the final manuscript.





\begin{appendices}
\renewcommand{\baselinestretch}{0.95} 
\section{An example of a user sample in our dataset.}

\renewcommand{\arraystretch}{1.2}
\begin{table}[h]
    \centering
    \begin{tabular}{|l|l|}
        \hline
        \multicolumn{2}{|c|}{\textbf{User Information}} \\
        \hline
        \textbf{Field} & \textbf{Information} \\
        \hline
        Name & User0 \\
        Age & 13 \\
        Gender & Female \\
        Ethnicity & Other \\
        Race & White \\
        Depression & No \\
        Anxiety & No \\
        Self Esteem Issues & Yes \\
        Bullying History & No \\
        Disciplinary Issues & Yes \\
        Substance Abuse & No \\
        Internet Usage & Weekly use less than 1 hour \\
        \hline
    \end{tabular}
 
    \centering
    \begin{tabular}{|c|c|p{7cm}|}
        \hline
        \multicolumn{3}{|c|}{\textbf{Cyberbullying Messages}} \\
        \hline
        \textbf{Sender} & \textbf{Bullying?} & \textbf{Message} \\
        \hline
        14 & B & TF???? EAT SOMETHING!!!!!!!!!!!!!!! \\
        159 & B & You should eat! \\
        66 & B & äöñNot all men-äöñ BITCH shut up !! Not all sharks bite ppl but Iäö'm still gonna be fuckin scared of ALL OF THEM bc they COULD. \\
        66 & B & I hope your 99 kids are fine \\
        144 & B & lmao how tf is this a costume for \$30 u can b a dumbass bitch and get ya ass beat by me fuck spirit \texttt{https://t.co/42jCKs1q7m} \\
        144 & B & Her skin is unique and beautiful, but she seems too thin. \\
        144 & B & Eat few burritos \\
        144 & B & Those Are Bones. Need to eat \\
        85 & B & i want to die in that \texttt{https://t.co/NqA4cy6fKB} \\
        85 & B &\raggedright Mino you've always been a salty bitch \texttt{https://t.co/9goStZXexJ} \\
        144 & B & Too thin \\
        144 & B & Too skinny \\
        48 & B & yeah fuck that guy with cancer who doesn’t have a job. Let him die! \\
        48 & B & you give up so easy, loser girl \\
        66 &B & "I know" \\
        66 & B & With \\
        66 & B & \raggedright Life of a little bitch \$10TUSD \texttt{https://t.co/kGD1ijDPIF} \\
        66 & B & PUT THE PUSSY ON A HIGH HORSE \\
        124 & B & I WANT \\
        124 & B & bro my sister been sleep since like 3 like uh bitch is you pregnant. \\
        \hline
    \end{tabular}
    \caption{Example of a user sample of the BullyBock dataset
}
    \label{appendix-table}
\end{table}

\end{appendices}

\bibliography{IEEEabrv,biblio_traps_dynamics}
\end{document}